\documentclass[letterpaper, 10 pt, conference]{ieeeconf}  
\usepackage{amsmath}
\usepackage{amsfonts}
\usepackage{multirow}
\usepackage{graphicx}
\usepackage{booktabs}
\usepackage{subcaption}
\usepackage{xcolor}
\usepackage{url}
\usepackage{hyperref}
\usepackage{bm}
\usepackage{bbm}
\usepackage{algorithm}
\usepackage{algpseudocode}

\IEEEoverridecommandlockouts                              

\title{\LARGE \bf
GOATS: Goal Sampling Adaptation for Scooping with Curriculum Reinforcement Learning
}

\author{Yaru Niu$^{1}$, Shiyu Jin$^{2,\dagger}$, Zeqing Zhang$^{2,3,\dagger}$, Jiacheng Zhu$^{1}$, Ding Zhao$^{1}$, Liangjun Zhang$^{2}$
\thanks{$^{1}$Carnegie Mellon University}%
\thanks{$^{2}$Robotics and Autonomous Driving Lab, Baidu Research, USA}%
\thanks{$^{3}$The University of Hong Kong}%
\thanks{$\dagger$Authors contributed equally to this work}
}

\begin{document}

\maketitle
\thispagestyle{empty}
\pagestyle{empty}

\begin{abstract}
In this work, we first formulate the problem of robotic water scooping using goal-conditioned reinforcement learning. This task is particularly challenging due to the complex dynamics of fluids and the need to achieve multi-modal goals. The policy is required to successfully reach both position goals and water amount goals, which leads to a large convoluted goal state space. To overcome these challenges, we introduce Goal Sampling Adaptation for Scooping (GOATS), a curriculum reinforcement learning method that can learn an effective and generalizable policy for robot scooping tasks. Specifically, we use a goal-factorized reward formulation and interpolate position goal distributions and amount goal distributions to create curriculum throughout the learning process. As a result, our proposed method can outperform the baselines in simulation and achieves $5.46\%$ and $8.71\%$ amount errors on bowl scooping and bucket scooping tasks, respectively, under 1000 variations of initial water states in the tank and a large goal state space. Besides being effective in simulation environments, our method can efficiently adapt to noisy real-robot water-scooping scenarios with diverse physical configurations and unseen settings, demonstrating superior efficacy and generalizability. The videos of this work are available on our project page: \textcolor{blue}{\href{https://sites.google.com/view/goatscooping}{https://sites.google.com/view/goatscooping}}. 


\end{abstract}

\section{Introduction}
\begin{figure}
    \centering
    \includegraphics[width=0.49\textwidth]{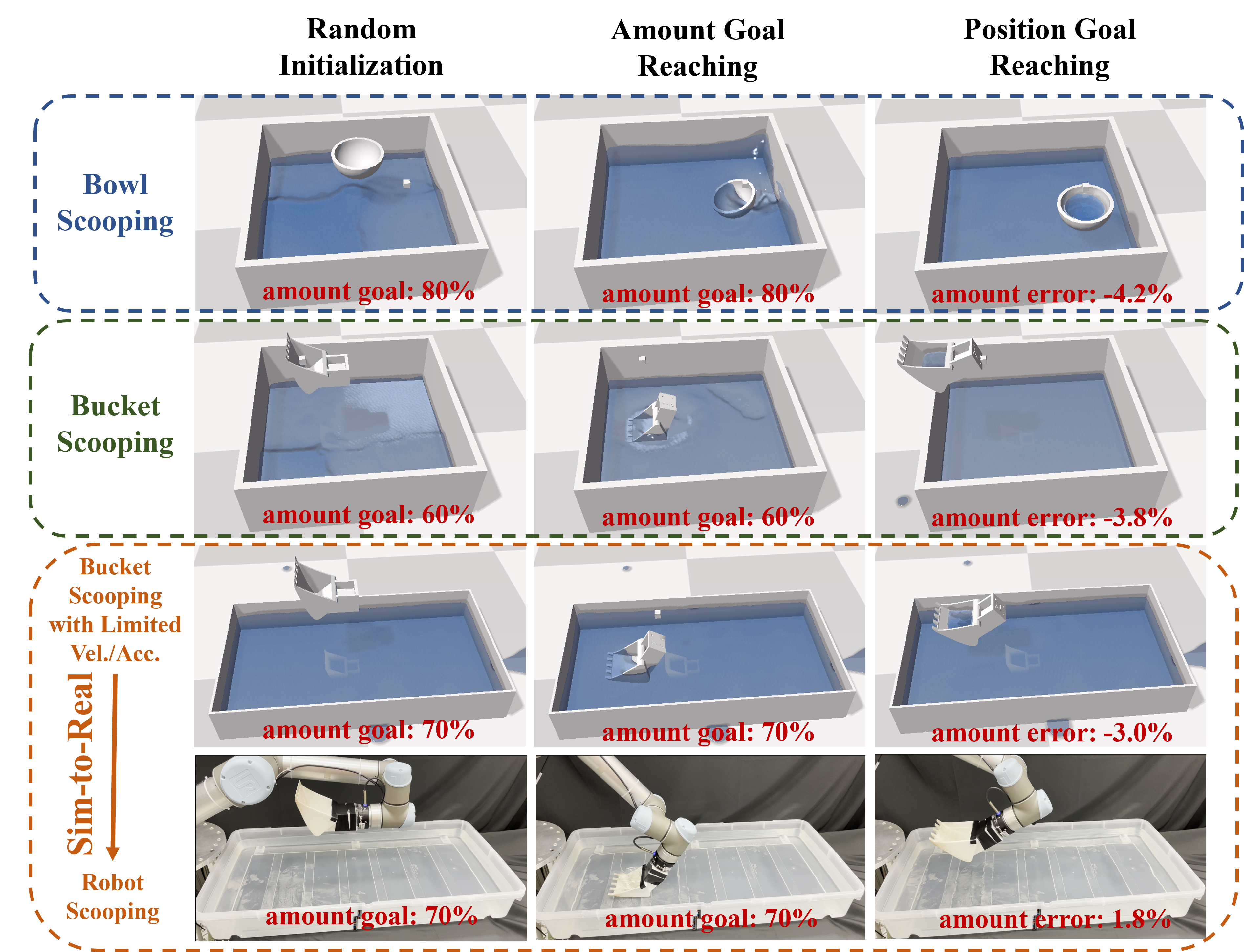}
    \caption{This figure depicts our goal-conditioned water scooping tasks. The task is randomly initialized over different water states (i.e., waterlines and fluctuations in the tank), different targeted water amounts and targeted positions (shown as a small white box). Our method can scoop the water to the targeted place with a small amount error using different containers in simulation, and can generalize well to real-robot scooping under various configurations.}
    \label{fig:intro}
    \vspace{-15pt}
\end{figure}

Scooping is an essential skill for humans to acquire in various aspects of life. We utilize tools such as spoons and shovels to efficiently scoop diverse types of materials, including fluids and granules. This skill can be applied to a wide range of tasks, from ladling soup and collecting peas at the dining table to excavating soil at a construction site. Despite extensive research in the field of robotic automation, only a limited number of studies have investigated autonomous scooping for robots, as demonstrated in \cite{schenck2017learning, wang2021learning, antonova2022rethinking, seita2022toolflownet}. Scooping poses a significant challenge due to the high-dimensional interaction between the end-effector and the dynamic materials. Furthermore, the problem of fluid or water scooping remains underexplored, with the complex dynamics of the fluid materials adding to the difficulty.
Performing goal-conditioned water scooping, such as scooping a specific amount of water from a tank and then transferring it to a designated location, would introduce even greater challenges.
Nevertheless, goal-conditioned water scooping holds significant potentials in both industrial and daily settings, facilitating downstream tasks like water transportation \cite{corl2020softgym}, water pouring \cite{tsuji2014high, chen2019accurate}, and caregiving \cite{erickson2020assistive, park2020active}.


In this work, we formulate the problem of goal-conditioned water scooping, and propose a goal sampling adaptation method for curriculum reinforcement learning method to solve long-horizon goal-conditioned scooping tasks. 
As shown in Figure \ref{fig:intro}, our proposed method can successfully scoop a specific amount of water from a water tank with small errors, and then reach a desired goal position with different containers in both simulations and physical robot settings. This task presents three main challenges. Firstly, it is a long-horizon task with a multi-modal goal state space which incorporates the position and water amount goals, so the policy is required to learn different types of motions to reach both goals, i.e., the container needs to first move downwards to scoop a targeted amount of water, and then lift to reach the desired position goal. Secondly, the initial state of the task is randomly initialized over different water states, and a large space of position goals and water amount goals. Thus, the policy needs to accommodate a wide range of high-dimensional random situations and has good generalizability. Thirdly, the water dynamics is complex, controlling the water amount of scooping under various changing conditions is nontrivial.




To this end,  our work is developed to solve these challenges, and we summarize our contributions as follows.
\begin{itemize}
    \item To the best of our knowledge, we are the first to formulate and benchmark the tasks of goal-conditioned water scooping with reinforcement learning, which would be beneficial to the robotics and robot learning community.
    \item We propose a goal-factorized reward formulation and a novel goal sampling adaptation method, GOATS, for efficient curriculum reinforcement learning on our water scooping tasks.
    \item Our proposed method achieves $5.46\%$ and $8.71\%$ amount errors on bowl scooping and bucket scooping tasks respectively, in simulation with 1000 variations of initial water states and a large desired goal space, and can control the amount error below $10\%$ in most cases on the real-robot scooping tasks. Our method can also generalize well to unseen tasks and maintain the same performance on two challenging unseen initial settings.
\end{itemize}

\section{Related Work}

\paragraph{Robotic Water Manipulation} 
Robotic water manipulation is a challenging problem as the dynamics of the water are complex to model. Many previous works have explored robotic water manipulation tasks, with a focus on pouring water from one container to another. Chen et al. propose to model water dynamics using a recurrent neural network and control pouring with model predictive control (MPC) \cite{chen2019accurate}. LaGrassa and Kroemer learn a model precondition where the dynamics models are accurate for pouring water \cite{lagrassaplanning}. The above methods rely on an accurate water dynamic model which is time-consuming to compute and may not be accurate. Other approaches reduce the burden of dynamic modeling with model-free methods \cite{guevara2017adaptable, rozo2013force}. Guevara et al. use a goal-based method and directly parameterize the objective over the space of actions that a robot can perform \cite{guevara2017adaptable}. Rozo et al. propose to learn force-torque traces of pouring skills from human demonstration using Gaussian mixture regression \cite{rozo2013force}. However, those model-free methods suffer from generalization problems if the scenarios change. With the development of physics simulation, some works evaluate the pouring pose \cite{wu2020can} and learn pouring RL policy \cite{do2018learning} in simulation and transfer it to the real world. 
Despite the abundance of research on pouring water, to the best of our knowledge, there has been no study conducted on water scooping, especially goal-conditioned water scooping.

\paragraph{Goal-Conditioned Learning for Deformable Object Manipulation}
Compared to rigid object manipulation, deformable object manipulation presents challenges due to complex dynamics, large state spaces, and diverse sensing configurations. 
Many works rely on heuristics, expert demonstrations, or dense supervised signals.
They mainly focus on cables \cite{jin2022robotic, wang2019learning, jin2019robust, wang2022offline}, fabrics \cite{jangir2020dynamic, canberk2022clothfunnels}, bags \cite{seita2021learning}, or granular materials \cite{zhang2023grains}. Jin et al. learn cable routing manipulation primitives based on a spatial representation from goal configurations \cite{jin2022robotic}, while the goal spatial representation can only be applied to linear objects.
Wang et al. generate a visual plan between the initial and the goal configurations for cable routing tasks via casual InfoGAN \cite{wang2019learning}, but the visual trajectory might be infeasible for liquids.
Jangir et al. study position goal reaching on cloth by combining reinforcement learning and behavior cloning \cite{jangir2020dynamic}, while the methods are only verified in simulation, and the expert demonstrations are difficult to generate within our task.
Canberk et al. propose a canonicalized-alignment objective to avoid local minima for cloth unfolding \cite{canberk2022clothfunnels}. Seita et al. propose to embed image-based goal-conditioning into Transporter Networks and manipulate 2D and 3D deformable objects to the desired states \cite{ seita2021learning}. These works rely on preset or parameterized action primitives, which are not suitable for our multi-modal goal-conditioned water scooping tasks.






\paragraph{Learning or Optimization for Scooping}
Various scooping tasks have been studied in both household and construction scenarios. In the household scenario, Schenck et al. learn to scoop pinto beans with a predictive model \cite{schenck2017learning}. Wang et al. learn generative models for scoop manipulation skills \cite{wang2021learning}.  Antonova et al. employ Bayesian Optimization with differentiable simulation to scoop a cube from the water in a tank and raise it to the highest possible position \cite{antonova2022rethinking}. ToolFlowNet uses behavior cloning from point clouds to perform water pouring and ball scooping tasks \cite{seita2022toolflownet}.
In the construction scenario, Jin et al. study the excavation of rigid objects with offline reinforcement learning \cite{jin2023learning}. Egli et al. learn an RL policy for soil excavation with an analytical soil model \cite{egli2022soil}. All of these works are developed with model-based learning or optimization, or depend on human demonstrations, which can be challenging for fluids with complex dynamics. Besides, none of the above focus on solving fluid/water scooping tasks. In this work, we formulate the problem of the goal-conditioned water scooping and provide a generalizable model-free RL solution for this task.



\section{Background}
In this paper, we formulate the water scooping task as a goal-conditioned reinforcement learning problem. In this section, we will introduce the basics of reinforcement learning and Hindsight Experience replay.

\subsection{Reinforcement Learning (RL)}
A Markov Decision Process (MDP) $\mathcal{M}$ is defined by a tuple $(\mathcal{S}, \mathcal{A}, p, r, \gamma, \rho_0)$. $\mathcal{S}$ is the state space, $\mathcal{A}$ is the action space, $p(s_{t+1}|s_t,a_t)$ is the probability of transitioning from state $s_t$ to state $s_{t+1}$, when applying an action $a_t$, $r(s_t,a_t)$ is the reward received by an agent for executing the action $a_t$ in state $s_t$, $\gamma\in [0,1]$ is the discount factor, and $\rho_0(s_0)$ is the initial state distribution. 
A policy $\pi(a_t|s_t)$ gives the probability of an agent taking action $a_t$ in state $s_t$. 
The goal of RL is to find the optimal policy, 
$\pi^*=\arg\max_\pi \mathbb{E}_{\tau\sim\rho_\pi}\left[\sum_{t=0}^T{\gamma^tr(s_t,a_t)}\right]$ to maximize cumulative discounted reward, where $\tau=( s_0,a_0,\cdots,s_T,a_T)$ is an agent's trajectory sampled from a distribution $\rho_\pi$ induced by policy $\pi$, initial state $\rho_0$ and the transition probability $p$. 
In this work, we employ Soft Actor-Critic (SAC) \cite{haarnoja2018soft} as our backbone RL algorithm. SAC is based on the maximum entropy RL framework, which augments the objective with the expected entropy of the policy over $\rho_\pi$ as $\mathbb{E}_{\tau\sim\rho_\pi}\left[\sum_{t=0}^T{\gamma^tr(s_t,a_t)}+\alpha\mathcal{H}(\pi(\cdot|s_t))\right]$, where $\alpha$ is the temperature parameter of the entropy term. In many cases, SAC can induce better exploration and learn more robust policies. The objective of the soft critic function parameterized by $\theta$ can be formulated as:
\begin{align}
    J_Q(\theta)=&\mathbb{E}_{(s_t, a_t) \sim\mathcal{D}}[\frac{1}{2}(Q_\theta(s_t, a_t)-r(s_t, a_t) \nonumber\\
    &-\gamma\mathbb{E}_{s_{t+1}\sim\mathcal{D}, a_{t+1}\sim\pi}[Q_{\bar\theta}(s_{t+1},a_{t+1}) \nonumber\\
    &-\alpha\log\pi_{\phi}(a_{t+1}|s_{t+1})])^2]
\end{align}
where $\bar\theta$ is the parameter of the target critic network, and $\mathcal{D}$ is the replay buffer.
The objective of the soft actor parameterized by $\phi$ can be expressed as:
\begin{align}
    \label{eqn:sac-actor}
        J_\pi(\phi) = \mathbb{E}_{s_t \sim\mathcal{D}}[\mathbb{E}_{a_t \sim\pi_\phi}[&\alpha \log(\pi_\phi(a_t|s_t)) -  Q_\theta(s_t, a_t)]]
\end{align}

\subsection{Multi-Goal RL via Hindsight Experience Replay} \label{sec:her}
Hindsight Experience Replay (HER) is a method that can improve the sample efficiency of off-policy RL methods for environments with multiple goals and sparse rewards, thus alleviating the efforts in designing shaped rewards and dependencies on domain knowledge. Following the idea from Universal Value Function Approximators \cite{schaul2015universal}, the trained policy takes in both the state $s_t$ and the goal $g$ at time step $t$. Given a goal-based task, for the standard experience replay of a chosen off-policy RL method, the transition $(s_t||g, a_t, r_t, s_{t+1}||g)$ will be stored in the replay buffer, where $r_t$ is the output of a reward function $r(s_t, a_t, g)$. The key idea of HER is to revisit achieved hindsight goals and use them to construct new transitions used for training. A new transition corresponding to $(s_t, a_t, s_{t+1})$ can be represented as $(s_t||g', a_t, r_t', s_{t+1}||g')$, where $g'$ is a hindsight goal, and $r_t'=r(s_t, a_t, g')$. The achieved hindsight goals could be sampled from the states which come from the same episode as the transition $(s_t||g, a_t, r_t, s_{t+1}||g)$. HER can be regarded as an implicit curriculum learning method as it generates less challenging intermediate goals that the current policy is able to achieve.

\section{Methodology}
\subsection{Problem Formulation for Water Scooping}\label{sec:problem}
In this paper, we formulate the water scooping task as a goal-conditioned RL task. We aim to learn a policy parameterized by $\theta$, $\pi_{\theta}$, to control the container to scoop a specific amount of water in the tank and move to a targeted position above the tank. Thus we can conveniently apply this policy to various downstream tasks like water pouring and transportation. 

At the start of each episode, the initial water state in the tank, which encloses the water amount, dynamics, and the initial position of the container, is sampled from the environment's initial state distribution $\rho_0$. Meanwhile, the desired goal state $g_{\text{desired}}=\{g_{\text{desired}}^p, g_{\text{desired}}^a\}$ is sampled from a goal distribution $\rho_g$. Here, $g_{\text{desired}}^p\in \mathbb{R}^3$ is the desired position goal of the container in the workspace, and $g_{\text{desired}}^a\in[0\%, 100\%]$ is the desired water amount goal in the container. At time step $t$, the scooping policy $\pi_{\theta}$ will take in an observation $o_t$ from the environment, the desired goal state $g_{\text{desired}}$ (fixed through an episode), and an achieved goal state $g_{\text{achieved}}^t$, and output a policy distribution $\pi_{\theta}(o_t, g_{\text{desired}}, g_{\text{achieved}}^t)$. The achieved goal state $g_{\text{achieved}}^t=\{g_{\text{achieved}}^{p(t)}, g_{\text{achieved}}^{a(t)}\}$ is a mapping or a subspace vector from the current observation $o_t$. Here, it shares the same space as $g_{\text{desired}}$, and includes the real-time position of the container $g_{\text{achieved}}^{p(t)}$ and the water amount in the container $g_{\text{achieved}}^{a(t)}$. From the policy distribution $\pi_{\theta}(o_t, g_{\text{desired}}, g_{\text{achieved}}^t)$, an action $a_t$ can be sampled and executed. Then, the agent will obtain an observation for the next time step $o_{t+1}$, a newly achieved goal $g_{\text{achieved}}^{t+1}$, and a reward $r_t=r(g_{\text{achieved}}^{t+1}, g_{\text{desired}})$, where $r(\cdot)$ is the reward function of our defined water scooping task. As $g_{\text{achieved}}^{t+1}$ is a result of executing $a_t$ on $o_t$, so the reward function $r_t=r(g_{\text{achieved}}^{t+1}, g_{\text{desired}})$ is a transformed version of $r'(o_t, a_t, g_{\text{desired}})$, which is introduced in Section \ref{sec:her}. The objective of this problem is to maximize the expectation of the discounted total reward, similar to a general RL problem.
 
This task is challenging because the model needs to learn a policy to accommodate different high-dimensional initial water states and the desired goal states. What is more, the task requires long-horizon planning of the policy, i.e., the container needs to first move downwards to scoop water of a specific amount and then move to a targeted place.

\subsection{Goal-Factorized Reward Formulation}\label{sec:reward}
To enable the RL algorithm to learn a good policy for the defined goal-conditioned water scooping task, it is important to design a good reward function with regards to the achieved goal $g_{\text{achieved}}^{t+1}$ and the desired reward $g_{\text{desired}}$. For notation convenience, we also represent $g_{\text{achieved}}^{t+1}$ and $g_{\text{desired}}$ as the concatenated vectors of their position goal and amount goal vectors (e.g., $g_{\text{achieved}}^{p(t+1)}|| g_{\text{achieved}}^{a(t+1)}$ and $g_{\text{desired}}^p|| g_{\text{desired}}^a$), respectively. A binary and sparse reward formulation that does not require human engineering and shaping proposed in \cite{andrychowicz2017hindsight} is 
\begin{equation}
    r(g_{\text{desired}}, g_{\text{achieved}}^{t+1}) = -\mathbbm{1}(\|g_{\text{achieved}}^{t+1}-g_{\text{desired}}\|>\epsilon)
\end{equation}
where $\mathbbm{1}$ represents the indicator function, $\|\cdot\|$ represents a distance between two vectors, and $\epsilon$ is a tolerance value. This reward function means that a reward of 0 will be only obtained when the achieved goal and the desired goal are close enough, otherwise, there will be a penalty of $-1$. It has advantages over the shaped dense rewards when the shaped rewards are hard to define and there is a discrepancy between the optimized rewards and the true success condition of the task. It also encourages exploration because some behaviors might be inappropriate for the short term but will bring larger benefits in the long term. For example, to scoop the water, the container needs to first go down to get the water. The downward movements will get the container further away from its position goal above the water tank but it is essential for the long-term goal.

However, a fully-sparse reward is still not appropriate for our water scooping task, because the policy needs to achieve the water amount goal besides the position goal, and formulating a binary reward function for both types of goals will make it really hard to get any positive feedback during the training process. Meanwhile, we find using a simple-shaped reward function specified for the amount goal will not bring any discrepancy in reaching the water amount goal. This shaped reward can be simply defined by the difference between the current water amount and the desired amount. Therefore, we propose to factorize the goal states and construct a hierarchical reward function as follows:
\begin{align}
\begin{split}
    r(g_{\text{desired}}, g_{\text{achieved}}^{t+1})=& \mathbbm{1}(\|g_{\text{achieved}}^{p(t+1)}-g_{\text{desired}}^p\|\leq\epsilon)\times \\
    &(1-\|g_{\text{achieved}}^{a(t+1)}-g_{\text{desired}}^a\|)-1  
\end{split}
\end{align}

This reward function means that dense positive feedback will be produced when the container is close enough to the position goal. It takes advantage of both the binary sparse reward and the shaped (but simple) dense reward, and thus it can help to both position goal and amount goal reaching.

\subsection{Curriculum Learning via Factorized Goal Sampling Adaptation} \label{sec:goats}
\vspace{-7pt}
\begin{figure}[htbp]
    \centering
    \includegraphics[width=0.50\textwidth]{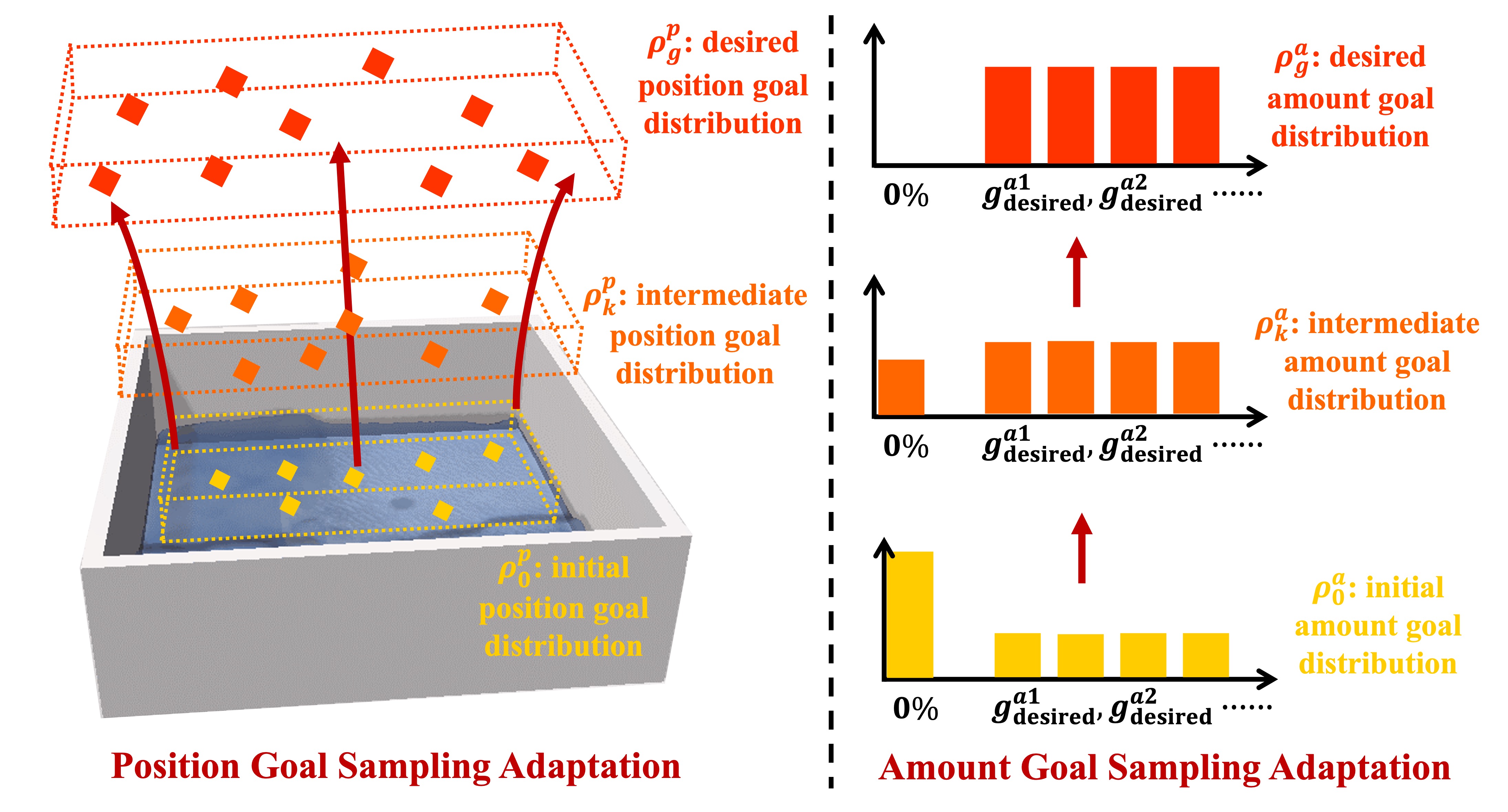}
    \caption{This figure demonstrates the process of position goal sampling adaptation and the amount goal sampling adaptation. Here, diamonds on the left are samples from the desired, interpolation, or initial distributions.}
    \label{fig:goats}
    \vspace{-9pt}
\end{figure}

As stated in \ref{sec:problem}, our desired goal state $g_{\text{desired}}$ subjects to a desired goal distribution $\rho_g$. Specifically, we represent it as $\rho_g=\{\rho_g^p, \rho_g^a\}$ which decomposes $\rho_g$ into a desired position goal distribution $\rho_g^p$ and a desired amount goal distribution $\rho_g^a$. To train a policy that can reach the goals from the desired goal distribution, a straightforward way is to directly sample goals from the goal distribution (i.e., task distribution) during training, like the approaches in \cite{andrychowicz2017hindsight, plappert2018multi, trott2019keeping}. However, our task is long-horizon and the initial state and the initial achieved goal distributions are not overlapped with, and sometimes can be far away from the desired goal distribution (e.g., the container with no water above the water tank v.s. the container with a targeted amount of water at a targeted place). This would lead to a total failure of our task. We display the corresponding results of SAC+HER in Section \ref{sec:sim_results}. Some other works on multi-goal RL develop explicit curriculum by sampling new goals that are in or close to the distribution of achieved goals to solve long-horizon problems \cite{warde2018unsupervised, pitis2020maximum, huang2022curriculum}, but they do not provide good solutions for our water scooping task with multi-modal goal distributions. 

In this work, we propose Goal Sampling Adaptation for Scooping (GOATS), which performs factorized goal sampling adaptation by generating intermediate position goal distributions and amount goal distributions. First, an initial position goal distribution, $\rho_0^p$, and an initial amount goal distribution, $\rho_0^a$, are selected to represent the initial state or initial achieved goal distribution of the task. 
Then the intermediate distributions, $\rho_k^p$ and $\rho_k^a$, are generated by interpolating between the initial goal distributions (e.g., $\rho_0^p$ and $\rho_0^a$) and the desired goal distribution (e.g., $\rho_g^p$ and $\rho_g^a$), respectively. Then the adaptive desired position goals $g_{\text{desired}}^{p(k)}$ and amount goals $g_{\text{desired}}^{a(k)}$ that represent simpler tasks can be sampled from $\rho_k^p$ and $\rho_k^a$.
A principled approach to measure the task distribution similarity is the 2-Wasserstein distance \cite{villani2009optimal}. Thus, the interpolations are the Wasserstein barycenters on a geodesic,
\begin{align}
\label{eq:w_barycenter1}
    \rho_k^p := \arg \min_{\rho'} (1 -k) W(\rho_0^p, \rho') + k W(\rho', \rho_g^p), \\
\label{eq:w_barycenter2}
    \rho_k^a := \arg \min_{\rho'} (1 -k) W(\rho_0^a, \rho') + k W(\rho', \rho_g^a),
\end{align}
where $W(\cdot, \cdot)$ denotes the Wasserstein distance between two distributions, $k\in[0, 1]$ is a temporal factor to indicate the procedure of the curriculum learning. 

As shown in Figure \ref{fig:goats}, in our scooping task, $\rho_0^p$ is a uniform distribution over positions on a cuboid region near the bottom of the water tank, $\rho_g^p$ is a uniform distribution over positions on a cuboid region that encloses the targeted region, $\rho_0^a$ is a distribution over the $0\%$ water amount, and equally distributed desired (discrete) amount goals, and $\rho_g^a$ is a uniform distribution over only the desired (discrete) amount goals. We select the initial position goal distribution at the bottom of the tank because the adaptive desired position goal $g_{\text{desired}}^{p(t)}$ can start from the locations near the water, which shortens the horizon between water scooping and position goal reaching.
We add $0\%$ water amount to $\rho_0^a$ because $0\%$ water amount will help the policy to learn to move to position goals in the workspace and enable the model to receive useful learning signal at the early training stage, which lowers the task difficulty. 
We do not interpolate distributions that include water amounts between $0\%$ and the desired amount goals.
This is because scooping a very small amount of water can sometimes be more difficult than scooping a larger amount of water; thus, it is not necessary to create too many goal amounts.
To this end, we explicitly parameterize the initial position goal distribution $\rho^p_0$ as a continuous uniform distribution $\mathcal{U}[\boldsymbol{a_0}, \boldsymbol{b_0}]$, where $\boldsymbol{a_0}$ and $\boldsymbol{b_0}$ are 3D vectors that represent the lower bound and the upper bound of the 3D space, respectively. Similarly, the desired position goal distribution $\rho^p_g$ can be parameterized as $\mathcal{U}[\boldsymbol{a_g}, \boldsymbol{b_g}]$. 
Then, the interpolation objectives in Equation (\ref{eq:w_barycenter1}) give closed-form solutions for the intermediate distributions as, 
\begin{align}
    \rho^p_k = \mathcal{U}[(1-k)\boldsymbol{a_0}+k \boldsymbol{a_g}, (1-k) \boldsymbol{b_0} + k \boldsymbol{b_g}].
\end{align}
For the intermediate amount goal distributions $\rho_k^a$, the solutions of Equation (\ref{eq:w_barycenter2}) are linear interpolations between the two discrete distributions, $\rho_0^a$ and $\rho_g^a$.
Given the explicit interpolation distributions, the intermediate desired position goals $g_{\text{desired}}^{p(k)}$ and amount goals $g_{\text{desired}}^{a(k)}$ can be adaptively sampled from $\rho_k^p$ and $\rho_k^a$ for training.

\subsection{Algorithm of GOATS}
GOATS combines the explicit curriculum learning procedure of goal sampling adaptation described in Section \ref{sec:goats}, and an implicit curriculum learning procedure HER, and provides an algorithm in Algorithm \ref{alg:GOATS}. We summarize it as follows. At the start of each episode, the temporal factor for curriculum learning $k$ will be first updated, and then $\rho_k^p$ and $\rho_k^a$ can be obtained by solving Equation (\ref{eq:w_barycenter1}) and (\ref{eq:w_barycenter2}) respectively. The adaptive desired position goal $g_{\text{desired}}^{p(k)}$ and  amount goal $g_{\text{desired}}^{a(k)}\sim\rho_k^a$ for this episode can be sampled from $\rho_k^p$ and $\rho_k^a$. Within each step of this episode, we will sample an action $a_t\sim \pi_{\theta}$, step the environment to get $o_{t+1}$ and $g_{\text{achieved}}^{t+1}$, and compute the reward $r_t$ with our proposed reward function in Section \ref{sec:reward}. Then we can update the replay buffer $R$ with this information, and update the policy $\theta$ using SAC and HER. Please note that here SAC can be replaced by any other off-policy RL algorithms.

\renewcommand{\algorithmicrequire}{\textbf{Input:}}
\renewcommand{\algorithmicensure}{\textbf{Output:}}
\begin{algorithm}
\caption{\small Goal Sampling Adaptation for Scooping (\textbf{GOATS}) with Curriculum Reinforcement Learning}\label{alg:GOATS}
\begin{algorithmic}
\Require Desired position goal distribution $\rho_g^p$, desired amount goal distribution $\rho_g^a$, initial position goal distribution $\rho_0^p$, initial amount goal distribution $\rho_0^a$, reward function $r(\cdot)$, initialized policy $\theta$, replay buffer $R$
\Ensure Learned policy $\pi_\theta$

\For{\texttt{each episode}}
\State Update temporal factor $k$ for curriculum learning 
\State Update adaptive desired goal distributions $\rho_k^p$ and $\rho_k^a$ \\
with $k$, $\rho_g^p$, $\rho_0^p$, $\rho_g^a$ and $\rho_0^a$ by solving Eq. (\ref{eq:w_barycenter1}, \ref{eq:w_barycenter2})
\State Sample adaptive desired goals \small $g_{\text{desired}}^{p(k)}\sim\rho_k^p$, $g_{\text{desired}}^{a(k)}\sim\rho_k^a$\normalsize
    \For{\texttt{each step} $t$}
    \State Sample action: $a_t\sim\pi_{\theta}(o_t, g_{\text{desired}}^k, g_{\text{achieved}}^t)$
    \State Step environment:$o_{t+1}\sim p(o_{t+1}|o_t, a_t)$
    \State Get $g_{\text{achieved}}^{t+1}$ from $o_{t+1}$
    \State Compute reward: $r_t = r(g_{\text{desired}}^k, g_{\text{achieved}}^{t+1})$
    \State Update replay buffer $R$
    \State Update policy $\theta$ via SAC, HER
    \EndFor
\EndFor

\end{algorithmic}
\end{algorithm}
\vspace{-12pt}

\section{Experiments}
In this section, we will first introduce the simulation environment settings we used for training the water manipulation policies. Then we summarize and analyze the results from our proposed method and baselines. At last, we demonstrate a real-robot water scooping experiment on a UR5 robot arm with a bucket container as its end effector.

\subsection{Simulation Setup}
We design the task and build our simulated scenarios based on SoftGym \cite{corl2020softgym}, a 3D simulator for deformable object manipulation by reinforcement learning. We test all the methods on two types of containers, including a bowl and a bucket, which have different shapes, sizes, and volumes, as shown in Figure \ref{fig:intro}. At the start of each episode, the initial water state is sampled from 1000 variations that are generated by dropping different numbers of water particles in the tank. This generation process produces different water fluctuations and initial waterlines that range from shallow levels to complete immersion of the containers. The duration of each episode in our task is set to 75 steps.

We use the state vector representation as the input to the network. The observation space is represented by positions, rotation, velocities, angular velocity of the container, water amount in the container, and estimated waterline in the tank. The goal state space includes the positions of the container and the water amount in the container. The actions for the scooping task include the accelerations and angular acceleration of the container. The radius of the bowl is 7.7cm, and the front length of the bucket is 11.7cm. For the bowl-scooping task, the position goal area is a cuboid from 27cm to 40cm above the ground (7cm to 20cm above the highest edge of the water tank), while for the bucket scooping, it is a cuboid from 37cm to 50cm (17cm to 30cm above the highest edge of the water tank) above the ground due to the size difference of the container. These are considerably large areas compared to the sizes of the containers and the length of the episode. The amount goals are $60\%$, $65\%$, $70\%$, $75\%$, $80\%$. We also compare the results of training on a single amount goal ($70\%$) and multiple amount goals in Section \ref{sec:sim_results}.

\subsection{Baselines in Simulation}
We provide the following six types of baselines to investigate the effectiveness of each component of GOATS.

\textbf{SAC} is a strong off-policy RL method that is widely used especially in environments with continuous action spaces. 

\textbf{SAC+HER} implements HER on SAC without any explicit curriculum learning.

\textbf{SAC+Universal Goal Sampling (GS)} does not apply adaptive goal distributions for SAC. Instead, it uses a fixed $\rho_k^p$ to cover a larger cuboid region that covers both the bottom region in the tank and the targeted region. It also fixes $\rho_k^a$ to sample the $0\%$ amount goal with a fixed probability of $50\%$.

\textbf{SAC+Partially Adaptive GS} uses the same $\{\rho_k^p\}$ as GOATS but fixes $\rho_k^a$ for SAC.

\textbf{SAC+HER+Universal GS} fixes $\rho_k^p$ and $\rho_k^a$ for SAC and HER.

\textbf{SAC+HER+Partially Adaptive GS} uses the same $\{\rho_k^p\}$ as GOATS but has a fixed $\rho_k^a$ for SAC and HER.

All the methods use the same hyperparameters for SAC and HER, and the models are trained on two desktops with Nvidia GeForce GTX 1080 Ti GPUs.

\subsection{Results in Simulation} \label{sec:sim_results}

\begin{figure*}[htp]
    \centering
    \includegraphics[width=0.92\textwidth]{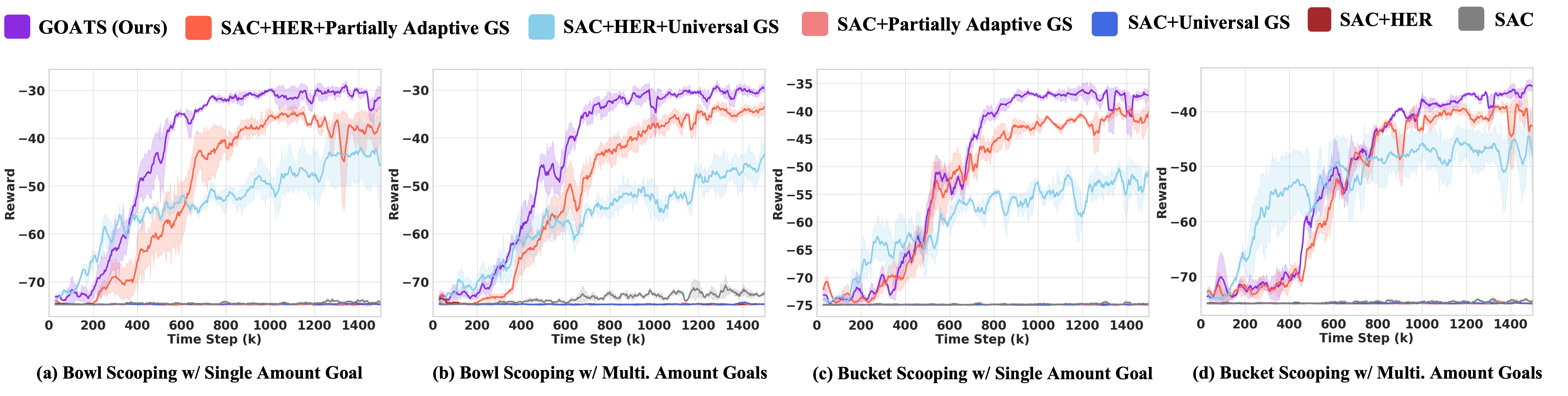}
    \vspace{-6pt}
    \caption{In this figure, we display the learning curves of different methods in the tasks of (a) bowl scoop with single water amount goal, (b) bowl scoop with multiple water amount goals, (c) bucket scoop with single water amount goal, and (d) bucket scoop with multiple water amount goals. The evaluation rewards under desired goals are averaged over three seeds, and the shaded region represents the standard error.}
    \label{fig:learning_curves}
    \vspace{-18pt}
\end{figure*}
We demonstrate our results from the simulation in Figure \ref{fig:learning_curves} and Table \ref{tab:sim_results}. The learning curves of four evaluated tasks from Figure \ref{fig:learning_curves} present similar trends. Our method, GOATS, can converge faster and achieve higher performance than all the baselines. The most competitive baseline is SAC+HER+Partially Adaptive GS, and it has very close performance to GOATS in the task of bucket scooping with multiple amount goals. The reward gaining of SAC+HER+Universal GS is larger than all other methods at the early training stage. The reason is that it directly samples the position goals from the entire space including the tank bottom and the position goal area, so it can apply the learned skills from the bottom to the position goal area at an early stage. All other methods totally fail in all evaluated tasks.

From the table results, we can tell that, besides reaching good rewards, GOATS can also achieve lower success water amount error than other methods (Table \ref{tab:sim_results}). This means that our method can successfully finish both the position goal and amount goal reaching tasks. In the task of bucket scooping with a single amount goal, SAC+HER+Partially Adaptive GS has lower rewards but shows fewer deviations from the amount goals. One possible reason is that the bucket might get a better amount of water without reaching the goal in some cases. Please note that the water amount error of SAC+HER in bucket scooping with multiple amount goals is higher than expected because one of the tested models is trained to bump into the tank and get some water but never reach the position goals. Comparing the amount errors between using a single amount goal and multiple amount goals on the same task, we can tell that training on multiple goals can be beneficial and improve the amount goal-reaching performance.

Compiling the results in both Fig. \ref{fig:learning_curves} and Table \ref{tab:sim_results}, SAC+HER+Partially Adaptive GS is always better than SAC+HER+Universal GS, from which we can conclude that performing position goal sampling adaptation is very helpful to our scooping tasks. Meanwhile, the performance discrepancy between GOATS and SAC+HER+Partially Adaptive GS shows the effectiveness of the amount of goal sampling adaption. The results here indicate that GOATS can accommodate complex water dynamics, different position and amount goals, and different types of containers in the water scooping task.

\begin{table*}[htp]
\centering
\caption{In this table, we display the performance of our proposed methods and the baselines in bowl scooping and bucket scooping tasks with a single water amount goal ($70\%$) or multiple water amount goals ($60\%, 65\%, 70\%, 75\%, 80\%$). The results are averaged over 3 seeds, and the standard errors are provided. For each seed, the reward is from the best evaluation reward during training. The corresponding model is then evaluated on 100 episodes, with randomly sampled initial waterlines, position goals, and water amount goals, to obtain the absolute amount error. The upper arrow means higher is better, and the down arrow means lower is better. The best performance of each task is marked in bold. } 
\label{tab:sim_results}
\resizebox{0.95\textwidth}{!}{%
\begin{tabular}{@{}ccccccccc@{}}
\toprule
\multirow{3}{*}{Method} & \multicolumn{4}{c|}{Bowl Scooping} & \multicolumn{4}{c}{Bucket Scooping} \\ 
\cmidrule(l){2-9} 
 & \multicolumn{2}{c|}{Single Amount Goal} & \multicolumn{2}{c|}{Multi. Amount Goals} & \multicolumn{2}{c|}{Single Amount Goal}  & \multicolumn{2}{c}{Multi. Amount Goals} \\ 
\cmidrule(l){2-9} 
 & \multicolumn{1}{c}{Reward $\uparrow$} & \multicolumn{1}{c|}{Amount Error$\downarrow$} & \multicolumn{1}{c}{Reward$\uparrow$} & \multicolumn{1}{c|}{Amount Error$\downarrow$} & \multicolumn{1}{c}{Reward$\uparrow$} & \multicolumn{1}{c|}{Amount Error$\downarrow$} & \multicolumn{1}{c}{Reward$\uparrow$} & \multicolumn{1}{c}{Amount Error$\downarrow$}
 \\\midrule
\multirow{1}{*}{SAC} & $-69.41 \pm 0.78$ & $69.60\% \pm 0.33\%$ & $-61.21 \pm 2.00$ & $71.02\% \pm 0.34\%$ & $-71.20 \pm 1.12$ & $69.99\% \pm 0.01\%$ & $-69.47 \pm 0.77$ & $71.00\% \pm 0.35\%$ \\ \midrule
\multirow{1}{*}{SAC+HER} & $-72.72 \pm 0.32$ & $67.28\% \pm 1.66\%$ & $-69.59 \pm 2.32$ & $63.36\% \pm 5.91\%$ & $-73.40 \pm 0.47$ & $52.35\% \pm 13.79\%$ & $-72.15 \pm 1.05$ & $55.76\% \pm 0.35\%$ \\ \midrule
\multirow{1}{*}{SAC+Universal GS} & $-71.7 \pm 0.69$ & $69.51\% \pm 0.40\%$ & $-72.05 \pm 0.41$ & $71.02\% \pm 0.34\%$ & $-72.96 \pm 0.65$ & $70.00\% \pm 0.00\%$ & $-71.48 \pm 1.28$ & $70.83\% \pm 0.43\%$ \\ \midrule
\multirow{1}{*}{SAC+Partially Adaptive GS} & $-72.89 \pm 0.59$ & $70.00\% \pm 0.00\%$ & $-71.87 \pm 0.18$ & $67.51\% \pm 2.23\%$ & $-73.73 \pm 0.24$ & $69.81\% \pm 0.15\%$ & $-73.14 \pm 1.01$ & $70.98\% \pm 0.35\%$ \\ \midrule
\multirow{1}{*}{SAC+HER+Universal GS} & $-36.45 \pm 4.41$ & $26.18\% \pm 14.33\%$ & $-37.88 \pm 2.48$ & $11.24\% \pm 2.51\%$ & $-42.48 \pm 1.04$ & $12.76\% \pm 2.60\%$ & $-37.32 \pm 1.24$ & $13.39\% \pm 0.69\%$ \\ \midrule
\multirow{1}{*}{SAC+HER+Partially Adaptive GS} & $-28.80 \pm 0.41$ & $8.54\% \pm 1.11\%$ & $-28.98 \pm 0.43$ & $7.43\% \pm 1.41\%$ &$-35.22 \pm 0.35$ & $\mathbf{9.61\%} \pm \mathbf{2.68\%}$ & $-33.12 \pm 0.60$ & $14.16\% \pm 3.16\%$ \\ \midrule
\multirow{1}{*}{\textbf{GOATS (Ours)}} & $\mathbf{-25.67} \pm \mathbf{0.32}$ & $\mathbf{5.93\%} \pm \mathbf{1.20\%}$ & $\mathbf{-25.77} \pm \mathbf{0.60}$ & $\mathbf{4.99\%} \pm \mathbf{0.37\%}$ & $\mathbf{-33.36} \pm \mathbf{0.69}$ & $9.97\% \pm 2.09\%$ & $\mathbf{-32.51} \pm \mathbf{0.61}$ & $\mathbf{7.45\%} \pm \mathbf{1.65\%}$ \\
\bottomrule
\end{tabular}%
}
\vspace{-10pt}
\end{table*}

\subsection{Sim-to-Real Transfer} To train a policy in our built scenarios that can be transferred to the real world and enable the robot to successfully perform the scooping task, we revise a few settings in simulation. First, we change the tank size in the simulation to match the real tank.
To ensure the scooping movements do not violate the physical limitations of the UR5, we constrain the acceleration and velocity ranges to $\frac{1}{5}$ of their original versions and lower the upper bound of the  position goal area to 33cm. Because the acceleration and velocity of the container are significantly decreased, we extend the training episode length to 110 steps to give the agent considerable time to finish the scooping task.

\subsection{Real-World Experiment}
\vspace{-9pt}
\begin{figure}[htbp]
    \centering
    \includegraphics[width=0.40\textwidth]{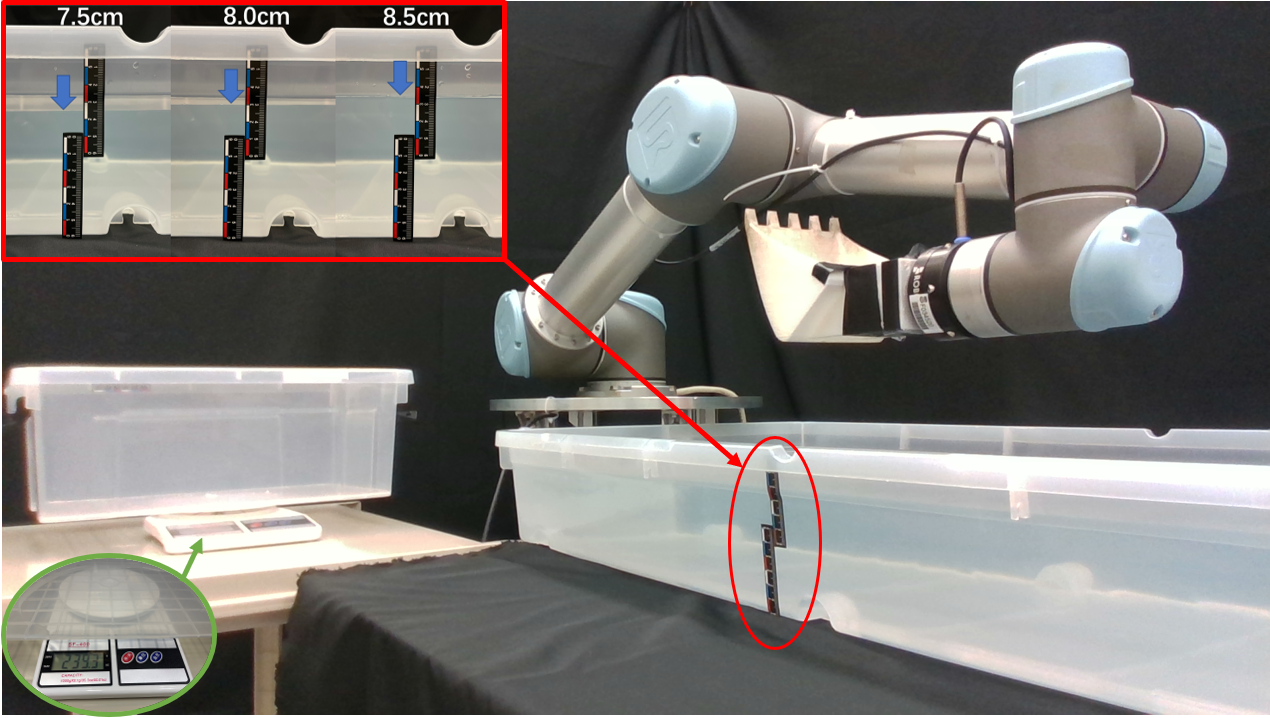}
    \caption{The figure shows our experiment setup, including the UR5 robot arm, 3D printed bucket, water container on the right-hand side, and a kitchen scale on the left-hand side. Note that the ruler line is stuck on the side wall of the container and we test three waterlines ($7.5, 8.0, 8.5$ cm).}
    \label{fig:exp_setting}
    \vspace{-9pt}
\end{figure}

\begin{figure*}[htp]
    \centering
    \includegraphics[width=0.97\textwidth]{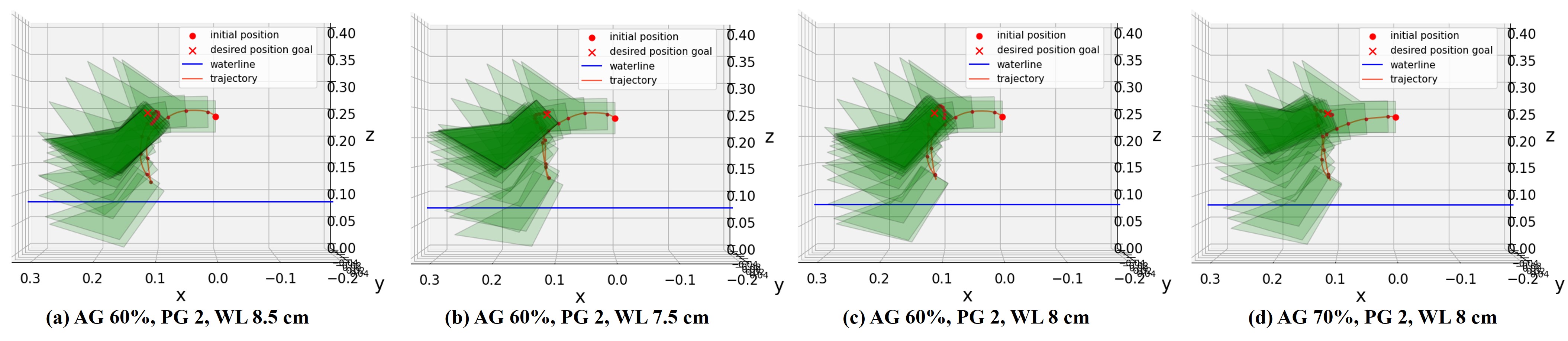}
    \caption{This figure depicts trajectories under different scooping conditions for the UR5 robot. Here, all initial positions are at $23$ cm above the ground. AG means the amount goal, PG means the position goal (all cases have the same PG), and WL means the initial waterline. The bucket dives deeper when the initial waterline is lower and the targeted amount is larger.}
    \label{fig:res_all_in_one}
\end{figure*}

\begin{table*}[htp]
\centering
\caption{In this table, we display the absolute water amount errors in both the sim-to-real simulation environment and the real robot environment using \textbf{a single trained model} by GOATS. Each value is an average over three tested position goals.} 
\label{tab:robot_exp}
\resizebox{0.97\textwidth}{!}{%
\begin{tabular}{@{}cccccccccccc@{}}
\toprule
\multirow{2}{*}{Initial Position}
 & \multirow{2}{*}{Waterline}
 &\multicolumn{2}{c|}{$60\%$ Amount Goal} & \multicolumn{2}{c|}{$65\%$ Amount Goal} & \multicolumn{2}{c|}{$70\%$ Amount Goal} & \multicolumn{2}{c|}{$75\%$ Amount Goal} & \multicolumn{2}{c}{$80\%$ Amount Goal}  \\ 
\cmidrule(l){3-12} 
 & (cm)
 & \multicolumn{1}{c}{Sim.} & \multicolumn{1}{c|}{Robot} & \multicolumn{1}{c}{Sim.} & \multicolumn{1}{c|}{Robot} & \multicolumn{1}{c}{Sim.} & \multicolumn{1}{c|}{Robot} & \multicolumn{1}{c}{Sim.} & \multicolumn{1}{c|}{Robot} & \multicolumn{1}{c}{Sim.} & \multicolumn{1}{c}{Robot}
 \\\midrule\midrule
\multirow{3}{*}{Height 23cm}
 & $7.5$ & $3.33\%\pm0.57\%$ & $1.66\%\pm0.75\%$ & $3.80\%\pm0.70\%$ & $6.58\%\pm3.06\%$ & $4.91\%\pm0.91\%$ & $11.27\%\pm1.75\%$ & $5.47\%\pm1.38\%$ & $14.05\%\pm1.44\%$ & $9.52\%\pm1.25\%$ & $13.55\%\pm1.24\%$ \\ \cmidrule(l){3-12} 
 & $8.0$ & $6.15\%\pm1.12\%$ & $4.39\%\pm2.94\%$ & $3.47\%\pm0.60\%$ & $2.67\%\pm0.98\%$ & $4.93\%\pm0.48\%$ & $4.95\%\pm2.62\%$ & $5.26\%\pm0.34\%$ & $9.58\%\pm0.83\%$ & $6.43\%\pm0.60\%$ & $11.47\%\pm1.72\%$ \\ \cmidrule(l){3-12} 
 & $8.5$ & $6.52\%\pm1.13\%$ & $5.64\%\pm0.83\%$ & $4.67\%\pm0.37\%$ & $3.74\%\pm1.51\%$ & $5.13\%\pm0.29\%$ & $5.31\%\pm2.79\%$ & $8.71\%\pm1.38\%$ & $11.08\%\pm1.85\%$ & $6.20\%\pm2.27\%$ & $9.71\%\pm1.57\%$ \\ \midrule
\multirow{3}{*}{Height 30cm} 
 & $7.5$ & $3.66\%\pm0.23\%$ & $1.76\%\pm0.18\%$ & $5.79\%\pm1.20\%$ & $7.83\%\pm1.95\%$ & $5.08\%\pm0.92\%$ & $11.43\%\pm1.73\%$ & $5.07\%\pm1.82\%$ & $11.51\%\pm2.47\%$ & $8.47\%\pm1.41\%$ & $14.23\%\pm1.18\%$ \\ \cmidrule(l){3-12} 
 & $8.0$ & $3.98\%\pm0.30\%$ & $4.78\%\pm1.81\%$ & $3.64\%\pm1.53\%$ & $1.83\%\pm1.19\%$ & $3.28\%\pm0.52\%$ & $5.07\%\pm2.14\%$ & $4.23\%\pm0.03\%$ & $8.64\%\pm1.38\%$ & $4.48\%\pm0.54\%$ & $11.35\%\pm2.80\%$ \\ \cmidrule(l){3-12} 
 (\textbf{Unseen in training})& $8.5$ & $6.53\%\pm0.91\%$ & $6.90\%\pm0.99\%$ & $5.17\%\pm1.25\%$ & $2.93\%\pm1.14\%$ & $4.25\%\pm0.75\%$ & $1.78\%\pm0.73\%$ & $6.10\%\pm1.19\%$ & $10.73\%\pm2.74\%$ & $6.53\%\pm0.97\%$ & $11.66\%\pm1.37\%$ \\\midrule
\multirow{3}{*}{Height 40cm}
 & $7.5$ & $4.31\%\pm2.31\%$ & $6.96\%\pm0.44\%$ & $5.21\%\pm0.99\%$ & $6.31\%\pm0.89\%$ & $3.92\%\pm0.50\%$ & $11.47\%\pm1.99\%$ & $4.49\%\pm0.81\%$ & $11.15\%\pm4.49\%$ & $7.06\%\pm0.90\%$ & $12.90\%\pm0.48\%$ \\ \cmidrule(l){3-12} 
 & $8.0$ & $3.94\%\pm0.56\%$ & $ 2.29\%\pm0.73\%$ & $3.06\%\pm1.34\%$ & $3.81\%\pm1.64\%$ & $4.13\%\pm1.01\%$ & $6.95\%\pm0.53\%$ & $4.47\%\pm0.54\%$ & $9.50\%\pm1.09\%$ & $4.81\%\pm0.84\%$ & $11.97\%\pm1.11\%$ \\ \cmidrule(l){3-12} 
 (\textbf{Unseen in training})& $8.5$ & $6.34\%\pm1.88\%$ & $7.90\%\pm1.97\%$ & $5.63\%\pm2.06\%$ & $3.85\%\pm0.52\%$ & $5.40\%\pm0.44\%$ & $5.11\%\pm3.04\%$ & $4.54\%\pm0.27\%$ & $10.75\%\pm1.80\%$ & $5.21\%\pm1.51\%$ & $10.31\%\pm5.21\%$ \\
\bottomrule
\end{tabular}%
}
\vspace{-13pt}
\end{table*}

To validate the practical scooping performance, we apply our trained policy to a real scenario as shown in Fig. \ref{fig:exp_setting}, where a 3D printed bucket of the same size as that in the simulation is mounted on a 7-DoF UR5. In addition, there is a long plastic box located on the black table storing water, and a kitchen scale on the white table is intended to measure scooped water from the bucket. 

Before the experiment, we adjust the waterline in the container by adding or reducing water to meet the prescribed height according to the ruler line on the side wall of the container. After the bucket returns to the initial pose, as presented in Fig. \ref{fig:exp_setting}, the UR5 could control the bucket to scoop water along the trajectory generated from waypoints given by the trained policy. Then, it will release the water into a box above the scale and the resulting water amount scooped by the bucket can be measured accordingly. 

We display the average absolute amount errors in both robot and (sim-to-real) simulation settings in Table \ref{tab:robot_exp}. We can find that when the amount goals (AG) are smaller than $75\%$, the amount errors of robot scooping are under $8\%$ in most cases, and the sim-to-real gap is small. With the increase of the targeted amount ($75\%$ and $80\%$), in simulation, the policy still produces scooping actions with small amount errors, while the errors on the robot get larger. One reason for this sim-to-real gap is that when the amount goal increases, the robot bucket usually dives deeper and sometimes it hits the bottom of the tank. Then it takes longer to lift the bucket than in the simulation, resulting in running-off of the water.
In addition, we demonstrate some trajectories from real robot experiments, as shown in Fig. \ref{fig:res_all_in_one}. In Fig. \ref{fig:res_all_in_one} (a) and Fig. \ref{fig:res_all_in_one} (b), we set the same water amount goal and position goal (PG) with two waterlines (WL), i.e., $8.5$ and $7.5$ cm. The trained policy can adaptively adjust the trajectory from the identical start point and PG to meet the same AG, and the bucket dives deeper when the waterline is lower. As a result, the amount errors on the robot of these two cases are $5.92\%$ and $3.07\%$, respectively. Furthermore, with the same WL, e.g., $8$cm in Fig. \ref{fig:res_all_in_one} (c) and Fig. \ref{fig:res_all_in_one} (d), different AGs are desired. It is shown that the bucket can dive deeper to get more water when the targeted amount is larger. The corresponding amount errors on the robot are $2.76\%$ and $-3.31\%$, respectively. Therefore, we can conclude that our trained policy can adapt to different water states in the tank and adjust scooping schemes to reach different amount goals on the physical robot.

\subsection{Generalization to Unseen Initial Bucket Positions}
In both the sim-to-real simulation environment and the physical robot environment, we directly apply the trained policy to unseen initial bucket positions in training without fine-tuning. The results are displayed in Table \ref{tab:robot_exp}. 
Compared to the performance on the in-distribution task (23cm), our method shows no evident performance drop on two more difficult out-of-distribution tasks (30cm, 40cm), and are surprisingly better at various amount goal and waterline settings.
This shows that GOATS has good generalizability and are empowered with the potential to achieve more complicated tasks when only training on simpler ones.

\section{Conclusion}

In this paper, we first formulate the goal-conditioned water scooping problem. This task is challenging due to the complex dynamics of fluid and multi-modal goal-reaching requirements. To tackle the challenges, we propose a goal-factorized reward formulation and a novel goal-sampling adaptation method for efficient curriculum reinforcement learning. We validate the effectiveness of our method on extensive water scooping experiments. Our method achieves better performance than baselines, on bowl scooping with an average amount error of $5.46\%$, and on bucket scooping with an average amount error of $8.71\%$ in simulation. In physical robot experiments, our method can adaptively generate trajectories for scooping and achieves amount errors lower than $10\%$ in most cases. What is more, our trained policy can generalize well to two more challenging unseen settings in both simulation and the physical world.


\addtolength{\textheight}{-12cm}   







\bibliographystyle{IEEEtran}
\bibliography{references}

\end{document}